\documentclass{article}

\usepackage{arxiv}
\usepackage{newtxmath}
\usepackage{balance}
\usepackage{microtype}
\usepackage{graphicx}
\usepackage{caption}
\usepackage{subcaption}
\usepackage{booktabs}
\usepackage{multirow}
\usepackage[table,xcdraw]{xcolor}
\usepackage{hyperref}
\usepackage{mathtools}
\usepackage{amsfonts}
\usepackage{amsmath}
\usepackage{nccmath}
\usepackage{pgfplots}
\usepackage{algorithm}
\usepackage{algorithmic}
\usepackage{lipsum}  
\newcommand{\link}[1]{{\color{blue}\href{#1}{#1}}}

\title{Data-Driven Reinforcement Learning for Virtual Character Animation Control}

\author{
  Vihanga Gamage \\
  School of Computer Science\\
  Technological University Dublin\\
  Dublin, Ireland \\
  \texttt{vihanga.gamage@tudublin.ie} \\
   \And
   Cathy Ennis \\
   School of Computer Science\\
   Technological University Dublin\\
   Dublin, Ireland \\
   \And
   Robert Ross \\
   School of Computer Science\\
   Technological University Dublin\\
   Dublin, Ireland \\

}

\begin{document}
\maketitle

\begin{abstract}
Virtual character animation control is a problem for which Reinforcement Learning (RL) is a viable approach. While current work have applied RL effectively to portray physics-based skills, social behaviours are challenging to design reward functions for, due to their lack of physical interaction with the world. On the other hand, data-driven implementations for these skills have been limited to supervised learning methods which require extensive training data and carry constraints on generalisability. In this paper, we propose RLAnimate, a novel data-driven deep RL approach to address this challenge, where we combine the strengths of RL together with an ability to learn from a motion dataset when creating agents. We formalise a mathematical structure for training agents by refining the conceptual roles of elements such as agents, environments, states and actions, in a way that leverages attributes of the character animation domain and model-based RL. An agent trained using our approach learns versatile animation dynamics to portray multiple behaviours, using an iterative RL training process, which becomes aware of valid behaviours via representations learnt from motion capture clips. We demonstrate, by training agents that portray realistic pointing and waving behaviours, that our approach requires a significantly lower training time, and substantially fewer sample episodes to be generated during training relative to state-of-the-art physics-based RL methods. Also, compared to existing supervised learning-based animation agents, RLAnimate needs a limited dataset of motion clips to generate representations of valid behaviours during training.

\end{abstract}

% keywords can be removed

\section{Introduction}
Virtual character animation control is an area of great interest as engaging human characters have great potential as mediums for interaction in games and other applications \cite{corrigan2016engagement,virtualcharactersmig2018}. While motion capture methods can be used to create animation, it can be time-consuming and expensive if required to capture a wider, varied range of behaviours. As a result, data-driven animation synthesis has been a problem that has drawn a great deal of interest.

When exploring novel solutions to data-driven animation control, neural network-based approaches have been a popular choice. Earlier work used supervised learning to create agents that output robust, varied animation, while learning from a limited set of reference motions \cite{holden2015learning,holden2017phase,zhang2018mode,datadrivengaze}. However, due to their heavy reliance on the training dataset, these approaches offer limited flexibility, and can lead to unpredictable outputs in conditions different to the training set. As a result, recent work has explored reinforcement learning (RL) to address this issue, with physics-based simulation being leveraged to implement character animation agents, such as DeepMimic \cite{peng2018deepmimic,liu2018learning}. Such methods rely heavily on interaction with physical surfaces and objects, as the feedback signals from the physics engine are required for agents to learn and function.

However, many applications using virtual humans require that they portray social, interactive behaviours, such as gestures, nods, exclamations or pointing. These behaviours do not elicit robust physics feedback signals, and therefore physics-simulation RL is not applicable. For portrayals of social behaviours to be effective, they need to be human-like, which poses a challenge from a modelling perspective. Taking pointing as an example, a function may be formulated that defines a successful pointing behaviour as one that returns to a starting pose after the direction of the index finger aligns with the vector to the target. Obtaining an optimal sequence out of the exponentially numerous possible trajectories is a problem without a straightforward solution, as the difference between a human-like portrayal and a more robotic one is difficult to define. Furthermore, for behaviours such as waving (in contrast to pointing) more dimensions would be required to articulate even a limited definition.

Recent work carried out using model-based RL has demonstrated that model-based agents can be trained to learn, within a compact latent state, dynamics required for agents to function robustly \cite{henaff2017model,hafner2019learning,sekar2020planning}. However, most state-of-the-art model-based RL approaches typically involve agents learning to solve tasks such as Cartpole and Walker from the DeepMind control suite \cite{tassa2018deepmind}. Animation control allowing for the portrayal of human-like behaviour is a more challenging task given that the dimensionality of the action space is much higher, and overall domain is more complex.

In this paper, we present RLAnimate, a data-driven RL approach to create model-based agents for virtual character animation control. We use a latent dynamics model to learn dynamics for animation and portrayed behaviour in a way that allows for agents to be robustly trained to generate animation portraying versatile, human-like behaviour. We formalise a framework for modelling the problem in a way that allows for a dual information state. This allows for leverage to learn two set of latent dynamics: one on the behaviour portrayed that is deterministic in nature, and the other for character animation dynamics universally applicable to any behaviour portrayed, which consists of deterministic and stochastic components. Agents are trained to output animation by self-generating a description for the next pose as per an objective signal, after which learned dynamics are applied to obtain latent representations used to calculate the rotations for the most optimal pose.

We summarise the contributions from our work as follows:
\begin{itemize}
  \item \textbf{Novel modelling framework for character animation}: we show that the mathematical structure we present that splits information into a behaviour portrayal-focused objective and a globally valid description, allows for the efficient training of a model-based agent to portray multiple behaviours.
  \item \textbf{Latent dynamics model for human-like animation}: by maintaining two latent spaces, the model learns to represent dynamics disentangling those for the behaviour portrayed, from those for the animation, which is the actuation of the behaviour portrayal. Our evaluation shows that this has a key impact on agent performance, and sample efficiency.
  \item \textbf{Training algorithm for RL animation agents informed by motion data}: modelling precise definitions for human-like behaviour, is challenging in terms of formulating a RL reward function. We implement a mechanism that allows for motion data to be used to inform the training process of representations for valid behaviour portrayal.
\end{itemize}

\section{Modelling human-like behaviour portrayal}
In our work, we are concerned with using RL to create virtual character animation control agents. An example of existing work applying RL to this problem is the work done by Peng et al. on DeepMimic learning physics-driven tasks \cite{peng2018deepmimic}. They combine an imitation objective with a task objective to train agents to output a range of physics-based character skills such as running and back-flips.

However, we are interested in agents generating animation that portray social behaviours such as gestures, that do not elicit requisite feedback from a physics engine to effectively train agents. Therefore, a novel approach that would enable RL agents to function, relying on signals that are ubiquitously present and applicable to animation regardless of behaviour, such as joint rotations and positions, is required.

\subsection{Modelling Reinforcement Learning Problems}
Problems that are to be resolved by the use of RL are generally described as a Markov Decision Process (MDP). Key components of a MDP are the state space $s\in S$, the set of actions $a\in A$, a state transition probability function $P \left ( s_{t+1} = s' | s_{t} = s, a_{t} = a \right )$ representing the probability of the successor state at $t + 1$ for a given action $a$ taken in state $s$ at time $t$, and a reward term $R_{a} \left ( s, s'\right )$ to represent the reward of a given action $a$ that leads to the transition from a state $s$ to a successor state $s'$. A RL algorithm then finds an optimal policy $\pi \left ( s \right )\rightarrow a$ which is a distribution over actions given states that maximises cumulative reward.

In using a MDP to define the problem, an assumption is made that the state space follows the Markovian property $P \left ( s_{t+ 1} | s_{t}\right ) = P \left ( s_{t+ 1} | s_{1},...,s_{t}\right )$, i.e.,  that the state at any given time depends only on its immediate predecessor. While assuming the Markov property is often needed to enable a theoretical structure that allows for convergence when learning optimal policies using RL algorithms, in some domains and settings, it may be more beneficial to explore alternate structures within which to apply RL algorithms. Model-free RL heavily relies on the assumption of the Markovian property which leads to the notion that all information required to obtain the ideal action is present in the representation of the state at the current time. An optimal model-free RL policy is a distribution that would allow for actions to be sampled per the state, to maximise reward without attempting to learn a model. Model-free RL is preferred in cases where a model is not available for the setting of the problem, or when not possible to learn an accurate model. Furthermore, the lack of reliance on a model makes model-free methods more adaptable to uncertainty or novelty during operation, relative to training. However, model-based RL offers many advantages such as sample efficiency if an accurate model for the dynamics relevant to the problem is present or can be learnt.

\subsection{Generating Human-like Animation}
With that in mind, we examine our problem domain of virtual human animation, where our goal is to generate optimal animation within a predictable range of behaviours, while the portrayals themselves are believable mimicry of human behaviour. To provide a brief overview of character animation, highlighting the key aspects required to understand the work presented in this paper, a virtual human character is a three-dimensional (3D) representation of a human, and animating a virtual character involves manipulating this 3D representation, typically consisting of a polygon mesh \cite{Schilinger2012}. This manipulation is enabled by the mesh being rigged with a hierarchical rigid-body structure, usually referred to as the skeleton, and consists of a configuration of connected joints \cite{par12}. The rotation of these joints results in animation.

The credit assignment problem is a prominent challenge in RL modelling, with a solution addressing the question of which actions taken over a period were most relevant to obtaining a particular reward \cite{kornfeld2020anatomical,weber2019credit}. In character animation, an output animation sequence needs to coherently portray a specific behaviour, and needs to be, out of millions of possible options, one that portrays a human-like version of the behaviour. An effective reward function would need to inform the training algorithm whether a particular shift in one trajectory makes the behaviour portrayed more accurate, more realistic, neither, or both. These two aspects, i.e., whether the overall behaviour is the right one, and whether this behaviour is human-like, are difficult to articulate mathematically.

\subsection{Mathematical Framework for Character Animation}
We address these challenges by following a model-based RL approach within a mathematical framework for modelling the virtual human animation tasks, that is refined to leverage attributes of the animation domain. While human understandable functions are not easy to define for behaviours such as gestures, motion capture data can be used to inform the agent based on latent definitions for ideal animation sequences. Compounded with the ability to learn accurate model for animation dynamics, this allows us to utilise the challenge of requiring consideration over a series of actions from preceding steps to portray human-like behaviour, as it would allow us to enhance the sample efficiency of the learnt models.

A goal of an animation agent is to generate an animation sequence $\left \{ a_{t} \right \}_{t=0}^{n}$, where $a_{t}$ represents the pose for each timestep for the total length of the sequence $n$. We split the state space into two: $s_{t} = o_{t}\cdot d_{t}$, with the state $s$ at a given time comprising of an objective $o$ and description $d$. The objective contains information on the behaviour that is to be portrayed, whereas the description contains information on the current pose of the character. This allows for animation dynamics being learnt in a behaviour agnostic manner.

Agents are trained to maximise the idealness $I$ of animation at each timestep, maintaining naturalness while successfully portraying the behaviour mandated via the objective state at each timestep. This overall training objective can be summarised as follows:

\begin{equation}
  E\left [ \sum_{t = 0}^{n}I_{t} \left ( a_{0},...,a_{t}, o_{t},d_{t}) \right ) \right ]
\end{equation}

For this split state space, the Bellman equations can be written as follows \cite{barron1989bellman}: the value of a state can be defined as $v_{\pi }\left ( s \right )$ defined as $\mathbb{E}_{\pi}\left ( I_{t+1} +  v_{\pi}\left ( S_{t+1} = O_{t+1}\cdot D_{t+1} \right ) | S_{t} = o \cdot d \right )$, and $q_{\pi }\left ( s,a \right )$ defined as $\mathbb{E}\left ( I_{t+1} + v_{\pi}\left ( S_{t+1} = O_{t+1}\cdot D_{t+1} \right ) | S_{t} = o \cdot d , A_{t} = a\right )$, which is the value of a pair of state and actions.

\section{Data-driven Reinforcement Learning}
In the previous section, we presented the mathematical framework we use to model human-like behaviour portrayal tasks in RLAnimate. The key challenge faced when training RL agents to model social behaviours is that a traditional RL environment can not be created, as it is difficult to generate formulae to assign rewards based on observations, particularly when the requirement for realism is considered. RLAnimate addresses this via agents learning dynamics for behaviour, and animation portraying these behaviours, which are then used to learn representations for valid, human-like behaviour. In this section, we describe the models learnt by an RLAnimate agent to do this, and the methodology by which they are trained.
\begin{figure}
  \includegraphics[width=\columnwidth]{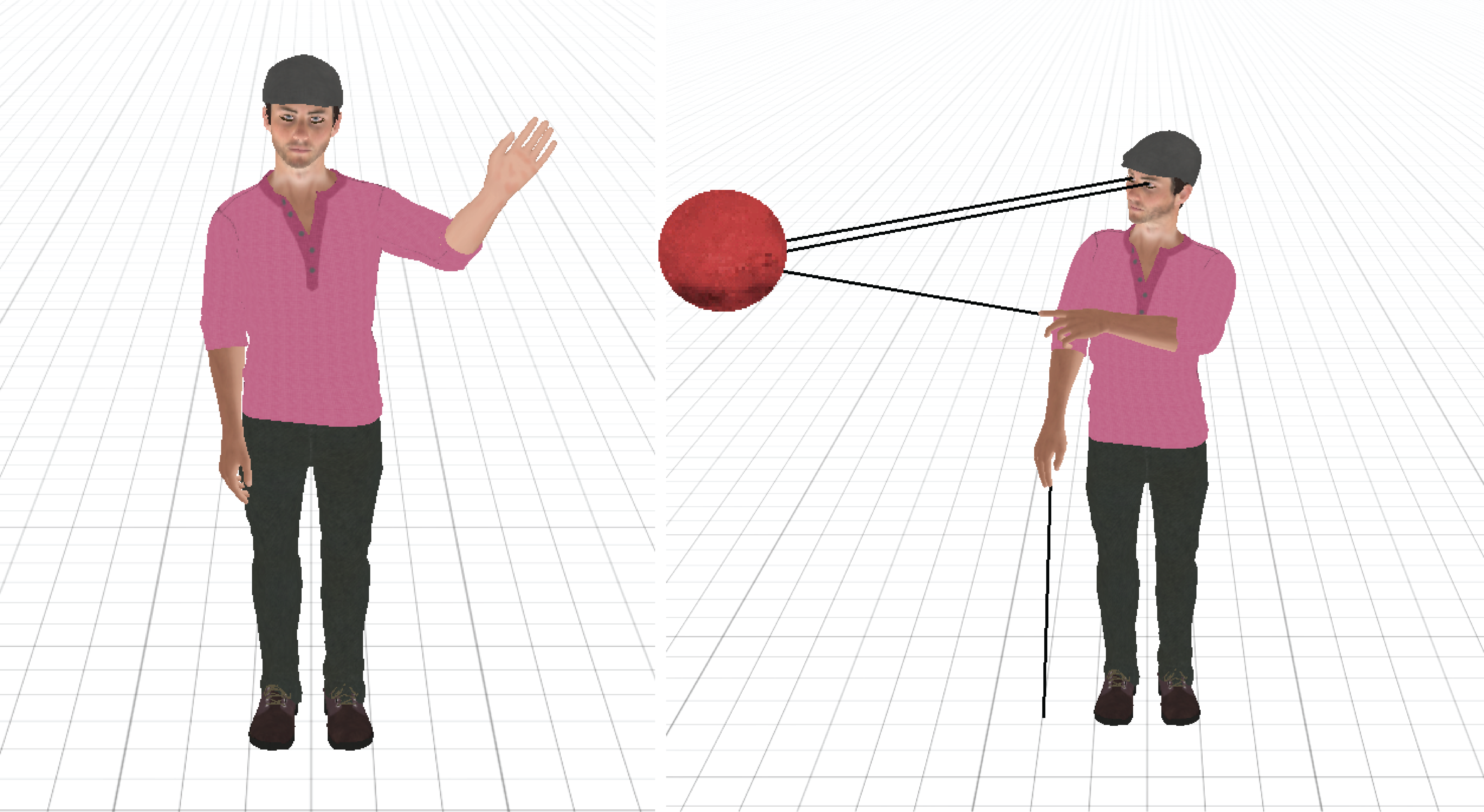}
  \caption{An overview of the waving (left), and pointing (right) behaviours used in this work. The vectors used in the description space are illustrated on the right.}
  \label{2_1_task_overview}
\end{figure}

\subsection{Agent Models}
\begin{figure}[ht]
\vskip 0.2in
\begin{center}
\centerline{\includegraphics[width=\columnwidth]{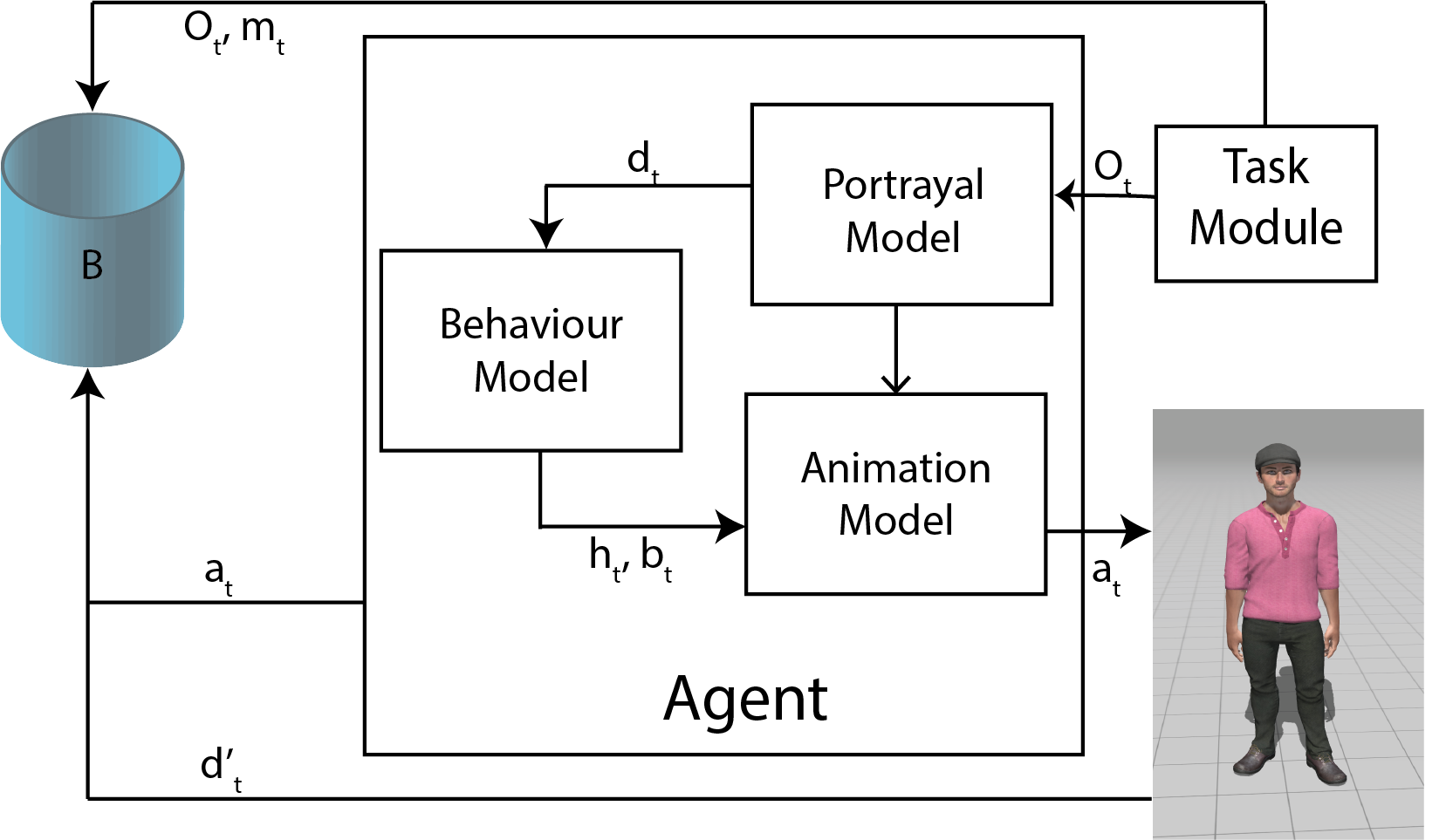}}
\caption{Overview of agent training and function. At each timestep, the agent recieves an objective signal $o_{t}$ from the task module in the environment. Using the portrayal model, the agent self-generates a corresponding ideal description $d_{t}$ for the next animation pose. This self-description in conjuction with the dynamics learnt allows an agent to obtain, from the behaviour model, latent states for task $h_{t}$ and behaviour $b_{t}$. These latent states are the used to generate the animation action $a_{t}$. The environment generates a real description signal which is stored in the sample buffer along with the action, objective and ideal animation $m_{t}$ per the relevant motion clip. }
\label{3_1_agent_overview}
\end{center}
\vskip -0.2in
\end{figure}

Figure \ref{3_1_agent_overview} presents an overview of our approach, within which agent function is a result of three co-operating models. We first present the design for these models before examining our data-driven RL algorithm in detail.

\subsubsection{Portrayal Model}
At each time step, the agent receives an objective state signal $o_{t}$ from the environment. The portrayal model $p(o_{t}|d_{t})$ is responsible for generating a description state $d_{t}$, based on the objective. The description state is a human-understandable vector containing information about the pose of the character, and the role of the portrayal model is to express a likely description that would satisfy the current objective state. Details regarding the description and objective state definitions are presented later in this section. To update models, RLAnimate collects episode data in a data buffer $B$. The task module that provides the objective signals passes $o_{t}$ to $B$ as well as the ideal motion $m_{t}$ to portray that objective obtained from the motion capture dataset.

\subsubsection{Behaviour Model}
The behaviour model is a latent state-space model that is responsible for learning latent animation dynamics. The design of this model was influenced by that of the recurrent state space model presented by Hafner et al. in Planet \cite{hafner2019learning}. We subscribe to their thinking that for robust dynamics to be learnt, the hidden states of the model needs to be split into deterministic and stochastic components. However, to address the challenges of portraying multiple behaviours, we incorporate several augmentations.

\begin{figure}[ht]
\vskip 0.2in
\begin{center}
\centerline{\includegraphics[width=\columnwidth]{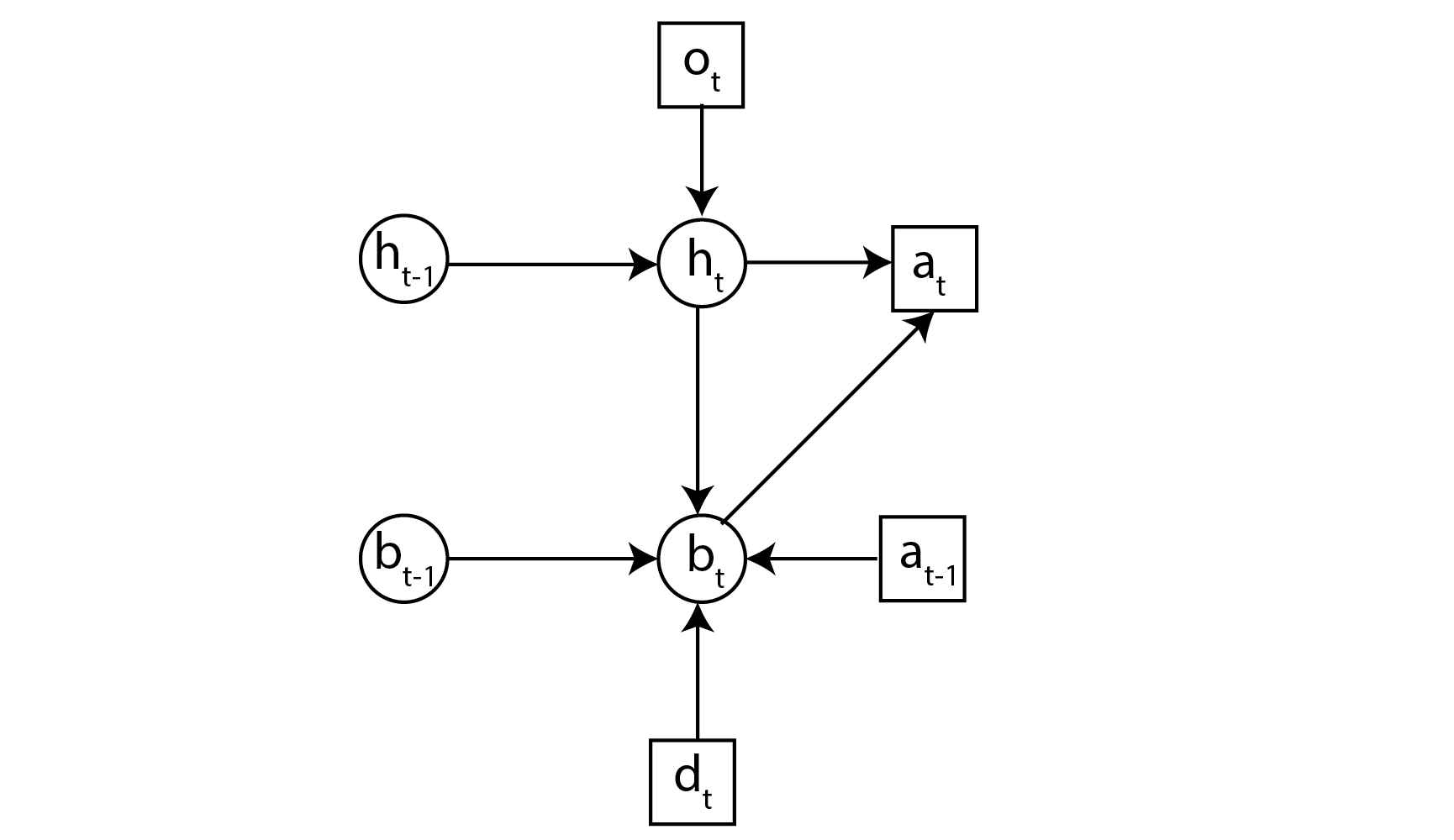}}
\caption{Latent dynamics learnt by RLAnimate agents. The deterministic task state $h_{t}$ is obtained as a recurrent neural network (RNN) $f_{1}(h_{t-1}, o_{t})$, and the deterministic component of the behaviour state as a second RNN $f_{2}(h_{t},b_{t-1},a_{t-1}$), which is used to generate the final behaviour state posterior $b_{t}$. $h_{t}$ and $b_{t}$ (computed as conditioned by $d_{t}$ generated by the portrayal model)is used by the animation model to generate the next animation action $a_{t}$.}
\label{3_2_dynamics_overview}
\end{center}
\vskip -0.2in
\end{figure}

Our behaviour model maintains a pair of latent states: the hidden task state $h_{t}$ and behaviour state $b_{t}$. The hidden task state is modelled as purely deterministic, whereas the behaviour state is split into deterministic and stochastic components. The manner in which the behaviour model applies learned dynamics to update these states can be summarised as follows:

\begin{fleqn}
\begin{equation}
\begin{alignedat}{2}
\text{Task state:}&\enspace & & h_{t} = f_{1}(h_{t-1}, o_{t})\\
\text{Behaviour state:}& & & b_{t} \sim p(b_{t}|f_{2}(h_{t},b_{t-1},a_{t-1}))
\end{alignedat}
\end{equation}
\end{fleqn}
An overview of the behaviour model design is shown in figure \ref{3_2_dynamics_overview}, and we will examine the model design and role of the dynamics learnt later in this section.

\subsubsection{Animation Model}
The animation model $p(a_{t}|h_{t},b_{t})$ receives the hidden task state $h_{t}$, and behaviour state posterior $b_{t}$ and outputs an action $a_{t}$ which are the rotations for the pose at that timestep of the animation sequence that are applied to the character model. The animation model output is parameterised as a Beta distribution, which is a class of continuous probability distributions that is defined within a bounded interval $[0,1]$, parameterised by two shape parameters $\alpha$ and $\beta$. We made this choice for output layer parameterisation, as Beta distribution-based policies have been shown to be more effective in continuous control reinforcement learning, by addressing issues caused by the mismatch between the infinite support of the Gaussian distribution and bounded controls \cite{chou17a}. The $a_{t}$ output by the agent is applied to the character, and stored in the episode buffer. When the animation is applied, the environment generates a real description $d'_{t}$ which is also sent to the episode buffer.

\subsection{Learning Animation Dynamics}
A core element of RLAnimate is that of leveraging the nature of character animation. With the problem of portraying realistic character animation comes several  complexities. One of which is the difference between realistic and unrealistic behaviour being ineffable in terms of defining a human-understandable function for computational purposes. This challenge is further enhanced by the nature of what character animation is portraying: human behaviour. Consider a portrayal where a human character points at a target straight ahead with the left arm. There are an exponentially large number of possible trajectories the arm could take, some more realistic than others.

However, the mechanics of the arm do not change, regardless of the target being pointed at, or even if the portrayal is of an entirely different behaviour such as waving. Hence, while human behaviour may very well be infinite in its complexity, our action space, which in this case is an approximation of the human body in the form of the joints used when modelling character animation, makes for a very structured medium for learning.

\subsubsection{Objectives and Descriptions}
This consistently structured action space of animation affords us leverage in order to address the challenges posed by the complexity of portraying human-like behaviours. To enable this, we provide input to the agents splitting the states into objectives and descriptions, both of which are human understandable form.

\textbf{Objectives:}
The objective signal essentially tells the agent about the behaviour that it needs to portray. The configuration for objective we use in this work is as displayed in table \ref{3_1_objectives}.
\begin{table}[t]
  \centering
\caption{Objetive configuration. t is the current step of the sequence, and N the total number of steps. Note that for attributes, the signal for waving contains a repetition of the mood variable, to account for the differences in sizes.}
\label{3_1_objectives}
\vskip 0.15in
\resizebox{0.45\textwidth}{!}{\begin{tabular}{c|c|c|c|c|c|c|c|}
\cline{2-8}
                                                                      & \cellcolor[HTML]{C0C0C0}Type & \multicolumn{2}{c|}{\cellcolor[HTML]{C0C0C0}Arm}             & \multicolumn{3}{c|}{\cellcolor[HTML]{C0C0C0}Attributes}            & \cellcolor[HTML]{C0C0C0}Time \\ \hline
\multicolumn{1}{|c|}{\cellcolor[HTML]{C0C0C0}}                        &                              & \cellcolor[HTML]{EFEFEF}Left & \cellcolor[HTML]{EFEFEF}Right & \multicolumn{3}{c|}{\cellcolor[HTML]{EFEFEF}Unit vector to target} &                              \\ \cline{3-7}
\multicolumn{1}{|c|}{\multirow{-2}{*}{\cellcolor[HTML]{C0C0C0}Point}} & \multirow{-2}{*}{1}          & 0                            & 1                             & x                     & y                    & z                   & \multirow{-2}{*}{t/N}        \\ \hline
\multicolumn{1}{|c|}{\cellcolor[HTML]{C0C0C0}}                        &                              & \cellcolor[HTML]{EFEFEF}Left & \cellcolor[HTML]{EFEFEF}Right & \multicolumn{3}{c|}{\cellcolor[HTML]{EFEFEF}Exaggeration}          &                              \\ \cline{3-7}
\multicolumn{1}{|c|}{\multirow{-2}{*}{\cellcolor[HTML]{C0C0C0}Wave}}  & \multirow{-2}{*}{2}          & 0                            & 1                             & 0 - 1                 & 0 - 1                & 0 - 1               & \multirow{-2}{*}{t/N}        \\ \hline
\end{tabular}}
\end{table}

\textbf{Description:}
Based in a geometric space that is constant regardless of the behaviour, the description signal describes the current state of the character. It is a combination of information on the effectors and joint positions as denoted in table \ref{3_2_description}.
\begin{table}[t]
\centering
\caption{Description configuration. The description consists of information about the virtual human character, constructed using the unit directional vectors for the effector articles and positions for joint articles. These vectors and positions are illustrated in figure \ref{2_1_task_overview}(b)}
\label{3_2_description}
\vskip 0.15in
\begin{tabular}{|
>{\columncolor[HTML]{C0C0C0}}l |l|l|}
\hline
Attribute       & \cellcolor[HTML]{C0C0C0}Articles                                                                                        & \cellcolor[HTML]{C0C0C0}Size \\ \hline
Effector Vector & \begin{tabular}[c]{@{}l@{}}left \& right eyes, \\ and index fingers.\end{tabular}                                       & 12                           \\ \hline
Joint Position  & \begin{tabular}[c]{@{}l@{}}left \& right collars,\\ shoulders, elbows,\\ wrists, and index finger\\ bases.\end{tabular} & 30                           \\ \hline
\end{tabular}
\end{table}

\subsubsection{Latent Dynamics}
The split objective and description states is the first step that enables us to create an approach to learn latent dynamics for character animation. Referring back to the behaviour model described earlier in this section, the duality is mirrored in the latent dynamics states. The task state $h_{t}$ is purely deterministic, which allows it to capture the dynamics concerning the objective. Using our earlier analogy, there might be a large number of ways to point forward or wave, but determining whether the right action is being portrayed, or whether the target being pointed at is accurate, does not require stochastic dynamics.

The behaviour state $b_{t}$ is a more complex latent state that consists of deterministic and stochastic components. It is updated by applying learned dynamics to the deterministic hidden task state, using the current animation pose. The deterministic task state captures the dynamics related to the objective state, which concerns information with regards to the variations in behaviours portrayed. As a result, the dynamics learned to update the behaviour state involve universally applicable information on animation.

Due to the non-linear nature of the model, the hidden behaviour state cannot be directly computed. We infer a behaviour state posterior using an encoder from the predicted description provided by the portrayal vector, expressed by equation \ref{conditioning}. Therefore, the behaviour state posterior can be considered as a representation of the most realistic animation poses for a given time step.
\begin{equation}
  q(b_{1:T}|d_{1:T},a_{1:T})=\prod_{t=1}^{T}q(b_{t}|f_{2}(h_{t},b_{t-1},a_{t-1}),d_{t})
  \label{conditioning}
\end{equation}
\begin{algorithm}[tb]
   \caption{RLAnimate}
   \label{algorithm}
\begin{algorithmic}[1]
\STATE Initialise episode buffer $B$.
\STATE Initialise models with random parameters $\theta$.
\WHILE{not converged}
\IF{training for waving and pointing}
\STATE draw at random an example $\left ( m_{t} \right )_{t=1}^{F}$ for each from motion dataset $M$
\ELSE
\STATE draw a single example $\left ( m_{t} \right )_{t=1}^{F}$ from motion dataset $M$
\ENDIF
\FOR{step t = 1..F}
\STATE Update deterministic task state $h_{t}$ from $o_{t}$
\STATE Infer behaviour posterior $q\left ( b_{t} |d_{t},a_{t-1} \right )$ conditioned on $d_{t}$ generated via the portrayal model
\STATE Generate $a_{t}$ and apply to character
\STATE Collect real description $d'_{t}$
\ENDFOR
\STATE $B \rightarrow B \cup \left \{ \left ( o_{t},a_{t},{d}'_{t},m_{t} \right )_{t=1}^{F} \right \}$
\FOR{model update step s = 1..T}
\STATE Draw episode chunks $\left \{ \left ( o_{t},a_{t},{d}'_{t},m_{t} \right )_{t=k}^{L+k} \right \}_{i=1}^{C}\sim B$ at random from data buffer
\STATE Compute loss $L\left ( \theta  \right )$ per (4)
\STATE Update model parameters $\theta \leftarrow \theta -\alpha \triangledown _{\theta }L\left ( \theta \right )$
\ENDFOR
\ENDWHILE
\end{algorithmic}
\end{algorithm}

\subsection{Training Agents}
In the previous sections, we presented how an RLAnimate agent uses latent dynamics to represent behaviour portrayals and generate animation. In addition to learning latent dynamics, to portray behaviour, an agent also needs to obtain an understanding of valid behaviour portrayals. As described in algorithm \ref{algorithm}, RLAnimate trains agents by generating rollout episodes of portrayals imitating motion clips provided, and iteratively updating the model parameters. To train agents, $C$ batches of episode chunks $\left \{ \left ( o_{t},a_{t},{d}'_{t},m_{t} \right )_{t=k}^{L+k} \right \}_{i=1}^{C}$ of length $L$ are drawn from the sample buffer where rollouts are collected in. The training objective seeking to maximise the ideality of animation consists of components $L1$ and $L2$ to learn latent dynamics for animation, and $L3$ to obtain an understanding of valid behaviour portrayals.
\begin{equation}
L\left ( \theta  \right ) 
\triangleq  E\left [ \sum_{t = 0}^{n}I_{t} \left ( a_{0},...,a_{t}, o_{t},d_{t}) \right ) \right ] 
\triangleq L1\left ( \theta  \right )  + L2\left ( \theta  \right )  + L3\left ( \theta  \right )
\end{equation}
The losses associated with learning the latent dynamics consist of the following description construction objective component $L1$ is calculated as the mean squared error, as expressed in equation \ref{loss_l1}, and KL divergence component ($L2$) based on the difference between the original description distribution and corresponding distribution obtained through the learned dynamics as expressed by equation \ref{loss_l2} \cite{hershey2007approximating}.
\begin{equation}\label{loss_l1}
L1\left ( \theta  \right )  = MSE\left [ p\left ( d_{t}|h_{t},b_{t} \right ) \right ]
\end{equation}
\begin{equation}\label{loss_l2}
L2\left ( \theta  \right )  = E\left [ q\left ( b_{t} |d_{t},a_{t-1} \right ) - p\left ( b_{t}|h_{t-1},o_{t-1},b_{t-1},a_{t-1} \right )) \right ]
\end{equation}
$L3$, as denoted by equation \ref{loss_l3} is a Huber loss minimising the difference between the original animation from the motion clip, and the animation generated by applying the dynamics learnt by the portrayal and behaviour models \cite{huber1992robust}. If the losses are over the threshold $\delta$, a linear function is used; otherwise, the function is quadratic.
\begin{equation}
  \label{loss_l3}
  L3\left ( \theta  \right )  = \left\{\begin{matrix}
  \frac{1}{2}\left ( M_{t} - f\left ( h_{t},b_{t} \right )^{2} \right ) & \text{for} \left | M_{t} - f\left ( h_{t},b_{t} \right ) \right |\leq \delta \\
  \delta \left | M_{t} - f\left ( h_{t},b_{t} \right ) \right | - \frac{1}{2}\delta^{2}& \text{otherwise.}
  \end{matrix}\right.
\end{equation}
The losses calculated with $L3$ are backpropagated through the rollout sequences, gradients flowing through the behaviour model to the portrayal model, allowing the agent to learn representations for valid behaviours in correspondence with the latent dynamics model.

\section{Experiments and Evaluation}
For our implementation, we use a game environment that builds on the the Panda3D engine \cite{goslin2004panda3d}. From Adobe Mixamo, we obtained 50 motion clips for pointing that covered a wide range of targets using both arms, as well as 50 motion clips for waving at various levels of exaggeration for each arm \cite{adobemixamo}. To evaluate trained agents, we held back a selection of 10 motion clips that included 6 pointing behaviours and 4 waving behaviours.

These testing clips included 3 clips each of a pointing portrayal using each arm, and 2 each for portrayals of waving with different levels of exaggeration to the training clips. A detailed list of the motion clips used for training and evaluation, as well as an overview of the technical implementation is available in the supplementary material. When evaluating agents, we measure successful behaviour portrayal by using a metric that calculates the difference between joints positions generated playing the agent output and reference clips, to obtain a score out of 100 using $100 - $total error$/$frames. The error per frame is obtained as expressed in equation \ref{errorpf}. The total error is calculated by adding a penalty term for per frame errors over 1, per equation \ref{errortotal}.
\begin{equation}
  \label{errorpf}
error_{t}=\sum_{j=0}^{J}\left (  \left | p_{x}^{j} -p_{x}^{j}{}' \right | + \left | p_{y}^{j} -p_{y}^{j}{}' \right | + \left | p_{z}^{j} -p_{z}^{j}{}' \right | \right )
\end{equation}

\begin{equation}
  \label{errortotal}
  \text{total error} = \sum_{t=0}^{T}error_{t}+ max\left \{ 0,\log_{1.01} error \right \}
\end{equation}
We also calculate a smoothness metric for final agent behaviour portrayals out of 100. Particularly with neural network-based approaches, data-driven animation output can contain artefacts that cause the motion to appear shaky. We use a Savitzky-Golay filter to obtain a smoothed version of the animation sequence, obtain a sum for the differences relative to the difference between the original motion clip and its smoothed version \cite{press1990savitzky}.

A video that contains a series of visual demonstrations and comparisons that emphasises the differences in performance and output quality of the agents and controls evaluated can be found at \link{https://virtualcharacters.github.io/links/ALA2021}. In our future work, we plan to carry out a perceptual evaluation using human participants. All our experiments were carried out using a workstation with an Intel Core i7-8750H 2.2GHz CPU and a Nvidia GeForce GTX 1070 GPU. We found the average throughput time for an RLAnimate agent rendered game to output animation was 0.0167 seconds per frame.
\begin{figure*}
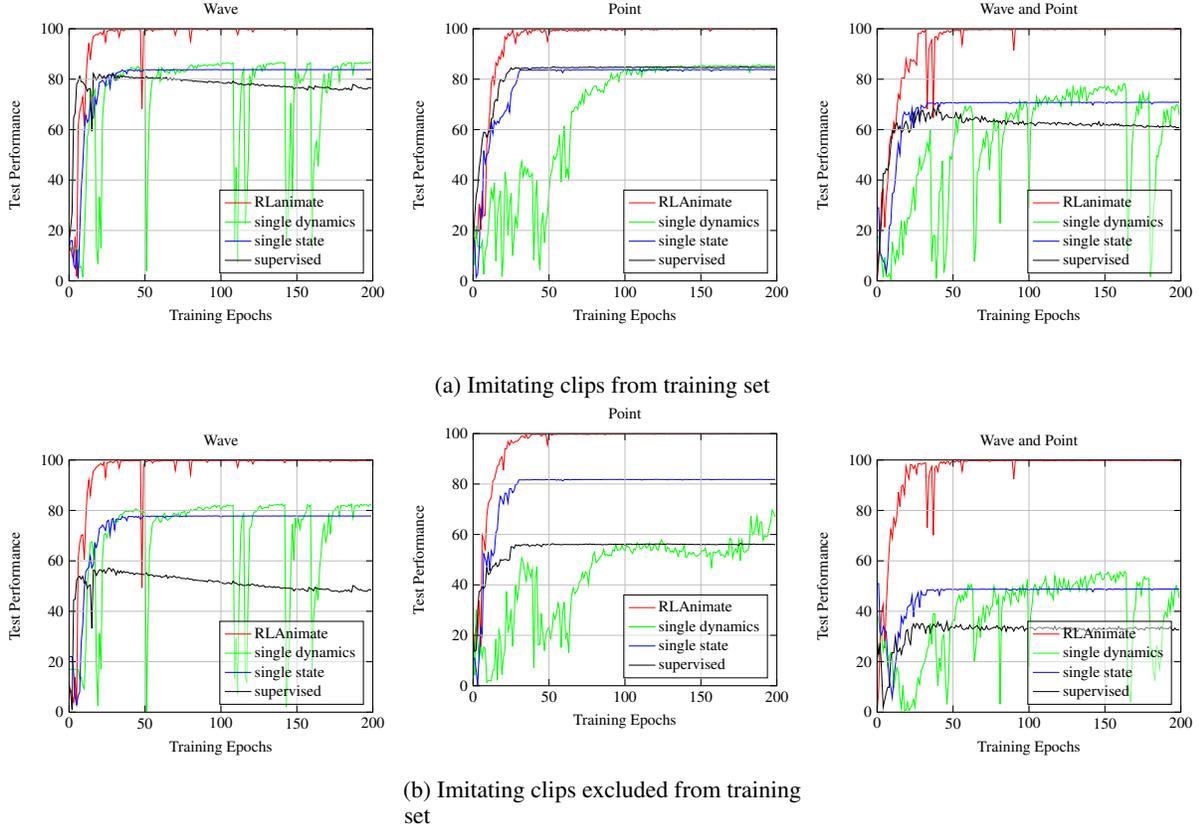

    \begin{subfigure}[b]{0.32\textwidth}
        \centering
        \resizebox{\linewidth}{!}{
        \input{plots/plot1.tex}
        }
        \label{fig:subfig8}
        \caption*{}
    \end{subfigure}
    \begin{subfigure}[b]{0.32\textwidth}
    \centering
        \resizebox{\linewidth}{!}{
        \input{plots/plot2.tex}
        }
        \label{fig:subfig9}
        \caption{Imitating clips from training set}
    \end{subfigure}
    \begin{subfigure}[b]{0.32\textwidth}
        \centering
        \resizebox{\linewidth}{!}{
        \input{plots/plot3.tex}
        }
        \label{fig:subfig10}
        \caption*{}
    \end{subfigure}

    \begin{subfigure}[b]{0.32\textwidth}
        \centering
        \resizebox{\linewidth}{!}{
        \input{plots/plot4.tex}
        }
        \label{fig:subfig8}
        \caption*{}
    \end{subfigure}
    \begin{subfigure}[b]{0.32\textwidth}
    \centering
        \resizebox{\linewidth}{!}{
        \input{plots/plot5.tex}
        }
        \label{fig:subfig9}
        \caption{Imitating clips excluded from training set}
    \end{subfigure}
    \begin{subfigure}[b]{0.32\textwidth}
        \centering
        \resizebox{\linewidth}{!}{
        \input{plots/plot6.tex}
        }
        \label{fig:subfig10}
        \caption*{}
    \end{subfigure}
\caption{Comparision of RLAnimate to other agent and model designs. The performance was measured over training epochs, with agents being trained for 50 epochs for single behaviour modes and 100 for agents learning to portray both pointing and waving. For the single behaviour modes, agents generated a single sample episode, whereas when training for both behaviours, agents generated two sample episodes, imitating a behaviour from each category.}
\label{results_plots}
\end{figure*}

\subsection{Ablation Studies}
We carried out a series of experiments to ascertain the effect of key RLAnimate elements. In addition to evaluating using the test set of motion clips withheld from training, we also scored agent behaviours using a set of 10 clips that were included in the training set.
\subsubsection{Spit observation and description Signals}
Figure \ref{results_plots} compares the performance of RLAnimate to a controller that provides the agent an input of the objective as a single state. The encoder used to parameterise the behaviour posterior (Equation (3)) is still trained using the description, but the self-description of the next ideal pose carried out using the portrayal model is eliminated. While the single state controller achieves a serviceable empirical score, it is notable that, particularly for waving, the animation generation contains significant artefacts, in the form of shakiness. The single state agent has the lowest smoothness measurement of all the agents we evaluated (Table \ref{results_table}), demonstrating that self-description by agents plays a key role in RLAniamte agents being able to learn animation dynamics effectively.

\subsubsection{Latent dynamics space}
We also compare the performance of RLAnimate to a control agent that used an alternate dynamics model that learnt animation dynamics using single latent space consisting of deterministic and stochastic components. This version is able to remain somewhat competitive relative to RLAnimate when portraying single behaviours, previously encountered in the test set. But when trained to portray either behaviour, there is a noticeable loss in performance. This demonstrates that using dual latent spaces to learn dynamics for tasks and animations separately allows agents to function more efficiently. Comparing the performance of an agent using the alternate dynamics model between imitating seen and unseen motions, while the difference is significant when trained to point or both, when waving, the difference in performance is much smaller, with the scores for the test set it had not encountered in training remaining within 4 points of imitating clips from the training set. This is due to the effect on animation due to the variation in degree of exaggeration being more predictable relative to the differences required to point at different targets. But, this agent is unable to maintain this performance when trained to perform either waving and pointing.

Additionally, the training for this agent appears significantly more unstable relative to the other agents. This affect is more pronounced and present throughout for waving and combined behaviours relative to pointing. This demonstrates further the benefit of using dual latent spaces to learn dynamics. Learning dynamics for animations via the descriptions universally, and separately learning dynamics for how objectives relate to ideal descriptions allows agents to function more effectively, particularly when portraying different behaviours. Even though this agent is able to learn to portray waving behaviours best, the mode of latent space does not allow for optimal dynamics to be captured, causing it to struggle. And while the agent seems to be able to learn dynamics in a more stable manner when trained to only point, it is unable to apply them effectively to portray unseen behaviours.

\subsubsection{Learning objective}
We also evaluated control agents trained with a completely supervised loss calculation. Their performance when imitating the test set was poor, containing a high amount of artefacts. As shown in Figure \ref{results_plots}, this agent performs better when portraying behaviours seen during training relative to the unseen testing set motions. This demonstrates that the objective we devise for RLAnimate directly enables it to generate dynamic animation, by optimising for learning optimal animation dynamics via one component and using its understanding of dynamics to optimise it to generate animation portraying ideal behaviours.

\subsection{Behaviour Portrayal Flexibility}
While RLAnimate enables agents to effectively vary the length of the behaviour portrayal relative to original clips, a key factor is whether these flexible animation outputs contain artefacts. We generated alternate episodes varying the length of the behaviour portrayals relative to the original behaviour, and measured the effect on mean smoothness of the motion trajectory. We found that RLAnimate agents are able to adjust without artefacts motion clip a range of 0.4x to 1.5x times the average length for that class of behaviour, before the artefacts become noticeable and the smoothness score dips below the 98.5\% threshold we identified. The supplementary video contains sequences of behaviours varying the length relative to the original motion length demonstrating this..
\begin{table*}[]
\centering
\caption{Final performance and smoothness of output generated imitating test set motions. After examining the output sequences,we identified 98.5\% as a threshold for the smoothness score after which artefacts become noticeable.}
\label{results_table}
\vskip 0.15in
\begin{tabular}{c|c|c|c|c|c|c|}
\cline{2-7}
                                            & \multicolumn{2}{c|}{\textbf{Wave}}       & \multicolumn{2}{c|}{\textbf{Point}}      & \multicolumn{2}{c|}{\textbf{Wave and Point}} \\ \hline
\multicolumn{1}{|c|}{\textbf{Method}}       & \textbf{Score} & \textbf{Smoothness} & \textbf{Score} & \textbf{Smoothness} & \textbf{Score}   & \textbf{Smoothness}   \\ \hline
\multicolumn{1}{|c|}{RLAnimate}             & 99.7                  & 98.5                & 99.8                  & 99.4                & 99.8                    & 99.1                 \\ \hline
\multicolumn{1}{|c|}{single state}          & 82.1                  & 96.2                & 66.9                  & 97.7                & 45.2                    & 97.0                 \\ \hline
\multicolumn{1}{|c|}{single dynamics space} & 77.8                  & 97.0                & 81.7                  & 97.1                & 48.8                    & 97.1                 \\ \hline
\multicolumn{1}{|c|}{supervised loss}       & 48.4                  & 96.8                & 56.0                  & 97.8                & 32.7                    & 97.3                 \\ \hline
\end{tabular}
\end{table*}

\section{Related Work}
Some of the earliest work applying neural networks for animation trained Convolutional Neural Network (CNN)-based models to portray cyclical behaviours such as human locomotion \cite{holden2015learning,holden2016deep,zhang2018mode}. However, performance of these models rely heavily on the training set, so they offer limited flexibility and ability to scale to portray a variety of behaviour types. Additionally, given the functions used to leverage the cyclical nature of behaviours resulted in increased computation during training, and final performance of agents required 30 hours.

More recent work has sought to generate animation portraying more dynamic behaviours by using physics simulation in conjunction with model-free reinforcement learning \cite{peng2017deeploco, peng2018deepmimic, liu2018learning}. But results so far have only demonstrated that physics-driven RL methods can be applied to tasks that feature interactions with physical surfaces and objects in the virtual environment. According to the authors, DeepMimic requires between 50 to 140 million sample episodes to be generated to train agents \cite{peng2018deepmimic}. RLAnimate requires episodes by a factor of about 0.5 million less; we believe this is due to RLAnimate being a model-based approach, requiring fewer sample episodes as an agent learns an effective model for animation dynamics. However, while DeepMimic agents are trained using a model-free RL algorithm, we point to their use of a physics engine at the core of the methodology. The role of the physics engine to provide feedback signals, is that of a model that governs the behaviour being portrayed. In RLAnimate, the latent dynamics model plays the same role, but the dynamics we learn depends on joint position and rotation-derived signals, which are ubiquitously applicable to character animation regardless of the behaviour being portrayed.

When exploring approaches to train agents portraying human-like behaviour, we chose model-based RL as an avenue of exploration due to the potential that would be afforded by learning latent dynamics applicable to multiple behaviours. Current work carried out has shown latent dynamics can be learnt to enable agents to function effectively \cite{lillicrap2015continuous, ha2018world, hafner2019learning}. We adopted a key conceptual element presented by Haffner et al. in PlaNet where they learnt latent dynamics models and used online planning to select actions \cite{hafner2019learning}. They also introduced a latent dynamics model which learnt both deterministic and stochastic components, allowing agents to robustly learn to make predictions about multiple futures.

But, in our work, we use a latent dynamics model with a further augmentation to learn dynamics as a pair of separate latent states: one component providing agents with representations for the most suitable behaviour to portray, and providing agents with a representation for an estimate of the most human-like portrayal of that behaviour. The former component is entirely deterministic, allowing it to maintain information precisely over the entire span of the behaviour. As the objective of this component is to provide an agent of an understanding of the different between pointing and waving, what each behaviour entails, and how to adjust to different speeds of portrayal, there is no need for stochastic dynamics. In fact, the rigidity by which it is mandated to capture dynamics gleaned from the objective state, enables the latter component to be more effective in learning animation dynamics applicable regardless of behaviour portrayed. We also opt to use a neural-network parameterized output method, as planning would not be efficient given the high dimensional nature of the action space.

\section{Discussion}
In this paper, we present RLAnimate, an approach for model-based animation control capable of portraying human-like behaviours. We observe that when applying RL for animation, the goal is the quality of the output animation, which needs to portray a finite, predictable range of behaviours, and we formalise a mathematical framework to model animation tasks along that line of thinking. We partition into objectives and descriptions, what typically would be the state space. During training, RLanimate agents learn a model to self-generate descriptions from the objective signal, and learn an advanced dynamics model to that maintains latent representations that can be used to obtain an animation sequence portraying natural human behaviour.

Our evaluation shows that RLAnimate agents are able to learn to portray different behaviours, using 0.5M x fewer sample episodes generated relative to physics-based model-free RL methods. And to inform the training algorithm of valid behaviour portrayal, we use a limited set of motion data that is significantly smaller in comparison to supervised learning-based methods.

Further, we note the sample-efficiency of the design of the modelling structure and latent dynamics models allowed for; particularly, the effectiveness by which agents could be trained to self-generate description signals of 54 dimensions from an objective signal of 6 dimensions to generate action signals with a dimension size of 45, that make up a human-like animation sequence. Accordingly, we believe approaches that seek to leverage the problem domain to learn more powerful models can be a promising avenue when applying model-based RL to a number of real-world problems and applications.

Research carried out examining human perception of character animation has shown that humans can be sensitive to the even minor differences. In our future work, we plan to ascertain the impact of small imperfections, and explore how agents can be influenced to avoid those pitfalls. We also plan to examine how RLAnimate can be applied to generate animation portraying a wider range of behaviours, in particular more complex portrayal of beat gestures and other complex behaviours required to interact with a user.

\bibliographystyle{unsrt}
\bibliography{references.bib}

\end{document}